\documentclass[11pt]{article}

\usepackage{acl2023}
\usepackage{times}

\usepackage{comment}
\usepackage[utf8]{inputenc} 
\usepackage[T1]{fontenc}    
\usepackage{hyperref}       
\usepackage{url}            
\usepackage{booktabs}       
\usepackage{amsfonts}       
\usepackage{nicefrac}       
\usepackage{microtype}
\usepackage{xcolor} 
\usepackage{color}
\usepackage{amsmath}
\usepackage{lipsum}
\usepackage{todonotes}
\usepackage{multicol}
\usepackage{enumitem}
\usepackage{graphicx}
\usepackage{balance}  
\usepackage{amsmath}
\usepackage[linesnumbered,ruled,vlined]{algorithm2e}
\usepackage{algpseudocode}
\usepackage{latexsym}
\usepackage{multirow}
\usepackage{multicol}
\usepackage{color}
\usepackage{dashrule}
\usepackage{extarrows}
\usepackage{dsfont}
\usepackage{centernot}
\usepackage{hyperref}
\usepackage{natbib}
\usepackage{fancybox}
\usepackage[utf8]{inputenc} 
\usepackage[T1]{fontenc}    
\usepackage{hyperref}       
\usepackage{url}            
\usepackage{booktabs}       
\usepackage{amsfonts}       
\usepackage{amsmath}
\usepackage{nicefrac}       
\usepackage{microtype}      
\usepackage{xcolor} 
\usepackage{amsmath}
\usepackage{lipsum}
\usepackage{colortbl}

\usepackage{todonotes}
\usepackage{multirow}
\usepackage{xcolor} 
\usepackage{bbm}
\usepackage{mdframed}
\definecolor{navyblue}{RGB}{0,0,128} 
\definecolor{vennred}{RGB}{255,0,0} 
\usepackage{xcolor}

\usepackage{xcolor}

\DeclareMathOperator*{\argmax}{arg\,max}

\definecolor{nav}{RGB}{0,0,128}

\newcommand{\model}{\textsc{SC$^2$}\xspace}

\title{Predicting Text Preference Via Structured Comparative Reasoning}

\author{
Jing Nathan Yan$^1$\thanks{~~Work done during the internship at Google. E-mail: \texttt{jy858@cornell.edu}.}~,~Tianqi Liu$^2$,~Justin T. Chiu$^1$,~Jiaming Shen$^2$,~Zhen Qin$^2$,~Yue Yu$^3$$^*$,~Yao Zhao$^2$, \\ \textbf{Charu Lakshmanan$^2$, Yair Kurzion$^2$, Alexander M. Rush$^1$, Jialu Liu$^2$,  Michael Bendersky$^2$}\\\\
\vspace{3pt}
$^1$\:Cornell University, 
$^2$\:Google,
$^3$\:Georgia Institute of Technology  \\
}

\begin{document}

\maketitle

\begin{abstract}
Comparative reasoning plays a crucial role in predicting text preferences; however, large language models (LLMs) often demonstrate inconsistencies in their reasoning, leading to incorrect preference predictions. While approaches like Chain-of-Thought improve accuracy in many settings, they struggle to consistently distinguish the similarities and differences of complex texts. We introduce \model, a model that prompts LLMs to predict text preferences by generating structured intermediate comparisons. \model begins by proposing aspects for comparison, followed by generating textual comparisons under each aspect. We select consistent comparisons with a pairwise comparator that ensures each comparison of a given aspect clearly distinguishes differences between texts, significantly reducing hallucination and improving consistency. Our empirical studies across various NLP tasks, including summarization, retrieval, and automatic rating, demonstrate that \model's enhanced performance in text preference prediction is significant.

\end{abstract}
\section{Introduction}
\begin{figure*}[t]
    \centering    
    \includegraphics[width=1\linewidth]{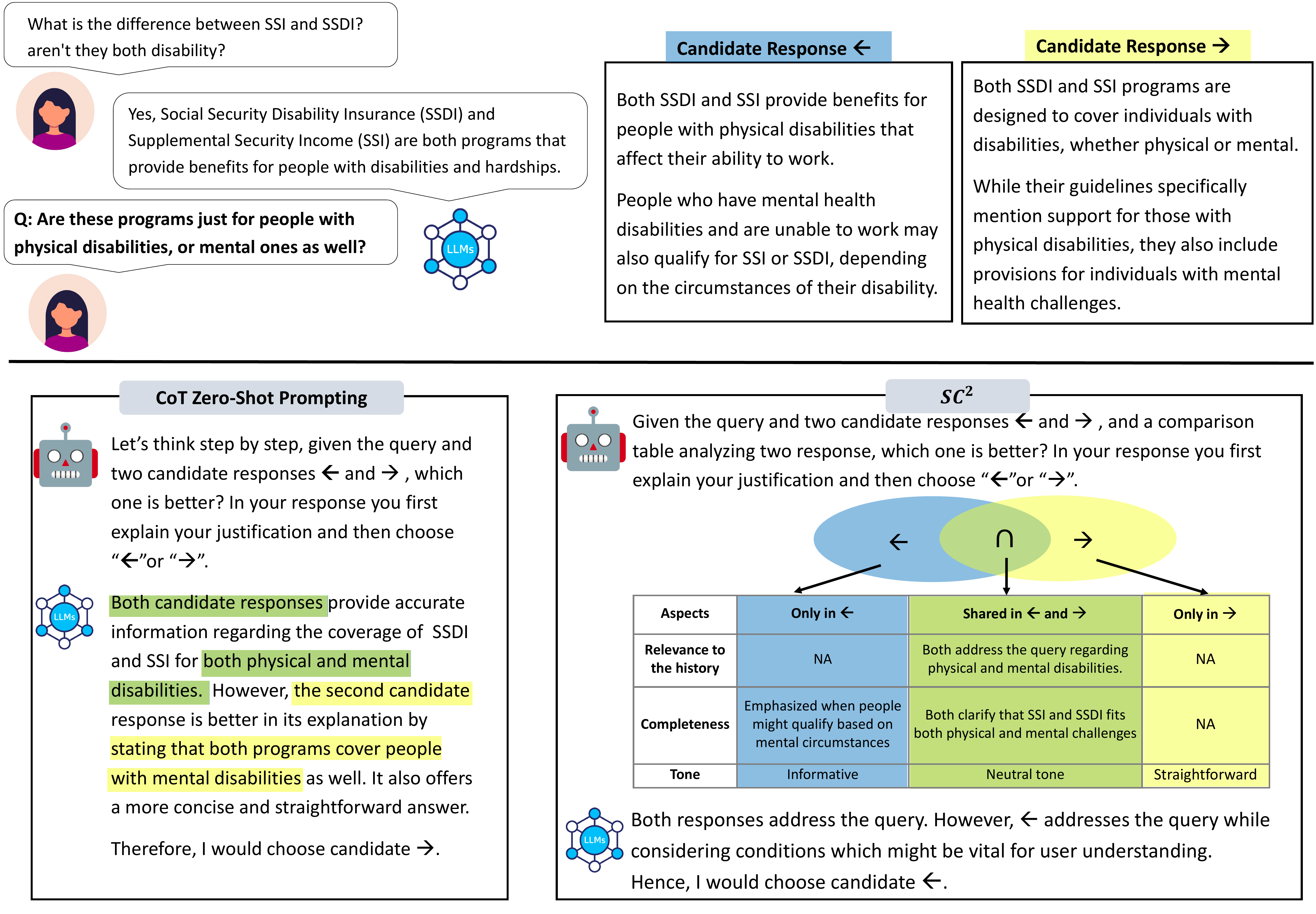}
\caption{An example illustrating the differences in predicting the text preference between CoT-Zero-Shot prompting and \model. The top portion shows a query between a human and a chatbot, along with two candidate responses $\leftarrow$ and $\rightarrow$. The table in the middle of the figure presents a structured intermediate representation produced by \model. Small phrases in the first row are \emph{aspects}. \emph{Comparisons} are entries not in the first column or row in the table. The Venn diagram visualizes the atomic comparisons for $\leftarrow$ and $\rightarrow$.}\vspace{-1em}\label{fig:intro:fig}
\end{figure*}

Comparative reasoning is crucial for predicting text preferences, as deciding the best out of a set of texts requires careful examination of the similarities and differences across the documents. Hence, comparative reasoning has been especially useful in NLP tasks such as text summarization~\citep{yang2023exploring, lee2023rlaif}, search ranking~\citep{qin2023large}, and automatic evaluation~\citep{adlakha2023evaluating}, where text preference prediction is a key step.

However, as corpora grow more dense and complex across domains, accurate comparative reasoning becomes increasingly challenging. Existing approaches rely on pretraining or fine-tuning models~\citep{yu2023pre, iso2021comparative} at the cost of massive human annotation and computation. With the emergence of large language models (LLMs)~\citep{gpt4,touvron2023llama, palm2, jiang2023mistral}, prompting approaches like Chain-of-Thought (CoT)~\citep{wei2022chain}  offer a promising solution for enhancing comparative reasoning. These approaches leverage LLMs' exceptional language generation capabilities without incurring significant overhead.

Nonetheless, LLMs exhibit arbitrary and erroneous outputs when prompted for comparative reasoning~\citep{adlakha2023evaluating}. Specifically, LLMs demonstrate inconsistency in their comparative analyses of texts. Figure~\ref{fig:intro:fig} (bottom left) provides an example of logically inconsistent LLM reasoning using zero-shot CoT prompting. The LLM's generated explanation initially describes a property as common to the text pair (highlighted in green), but later implies that the same property is a strength of just one of the documents (highlighted in yellow). This inconsistency in the LLM's comparative analysis leads to an incorrect prediction.

To address these challenges, we present \model, a \underline{S}tru\underline{C}tured \underline{C}omparative reasoning model that constructs an intermediate structured representation contrasting two text corpora for more accurate text preference prediction, as illustrated in Figure~\ref{fig:intro:fig}. First, \model proposes a set of aspects from text pairs to guide the comparison step. Second, \model generates textual comparisons for every aspect. We refer to aspects and comparisons together as intermediate structured representations. To improve the consistency of reasoning (e.g., a contrastive comparison of a aspect should not overlap with a common comparison), \model adopts approximate inference: \model samples multiple responses in generative process and uses a pairwise comparator to select the most consistent intermediate structured representation for final preference prediction.

We demonstrate the effectiveness of \model in improving text preference prediction across various tasks including text summarization~\citep{stiennon2020learning}, document retrieval~\citep{trecnews}, and helpfulness and harmlessness detection~\citep{bai2022training} with average $2.5$ and $7.0$ points gain over the top and bottom baselines, respectively.

Our analysis further confirms the effectiveness of \model without incurring expensive LLM usage, and ablation studies emphasize the importance of the pairwise comparator. Our extensive human evaluations also indicate that \model aids in interpretation and assists users in making decisions.

\section{Related Work}
\paragraph{Prompting Large Language Models}
LLMs have recently advanced the state-of-the-art performance across many NLP tasks~\citep{palm2,gpt4, chowdhery2022palm, touvron2023llama, touvron2023llama2, jiang2023mistral}. 
These LLMs have demonstrated the capability to provide chain-of-thought explanations that elucidate their reasoning processes~\citep{wei2022chain,kojima2022large}. However, the chain-of-thoughts generated by LLMs are often arbitrary or contradictory~\citep{Wang2022TowardsUC,turpin2023language,chen2023models,dhuliawala2023chain}, unfaithful to the facts~\cite{lyu2023faithful,lyu2024towards} or lacking robustness to rephrased questions. 
To mitigate these issues, several works aim to leverage consistency-based~\cite{wang2022self,zhou2022least}, or verification-based approaches~\cite{ling2023deductive,lyu2024towards} to improve the reasoning capacity of LLMs, yet the benefit of such additional techniques are still ambivalent~\cite{huang2023large}. 
Furthermore, all these advanced techniques still concentrate on processing raw-text inputs, thereby overlooking the integration of structural information.  Moreover, they lack implementations of explicit consistency constraints, which is crucial for maintaining logical coherence in generated outputs.

\paragraph{Comparative Reasoning and Summarization} 
Comparative reasoning involves comparing and contrasting different documents~\citep{yu2023pre},  which has applications for a broad range of 
NLP tasks including text ranking~\citep{jiang2023llm,qin2023large}, reward modeling~\citep{ouyang2022training,lee2023rlaif,zhang2024pla} and automatic text generation evaluation~\citep{liu2023gpteval}.  
Initial explorations focused on mining comparative content from text corpora~\citep{jindal2006identifying, li2011comparable}. More recent studies have developed models for generating comparative text, including generating arguments for answering comparative questions~\citep{chekalina2021better, amplayo2021aspect} and summarizing comparative opinions~\citep{iso2021comparative}. Additionally, \citet{zhong2022describing, zhong2023goal} prompt LLM to describe the differences between two text distributions in natural language and \citet{dunlap2023describing} further extends to discover differences given a set of images from ImageNet. 

One challenge of directly prompting LLMs for comparative reasoning
is that the input text often contains a mixture of diverse patterns. As such, it is crucial to incorporate fine-grained aspects to guide LLMs for generating more comprehensive summarizations~\cite{sun2023principle,xu2023knowledge,wu2023fine,yu2023large}.
Early works~\cite {lin2000automated,titov2008modeling} used clustering or topic modeling to identify aspects in documents.
\citet{lekhtman-etal-2021-dilbert} fine-tune a  pretrained language model for aspect extraction, which relies on manual labeling of comparative data. 
On the other hand, \citet{goyal2022news,yang2023exploring} leverage LLMs to perform summarization with the fixed aspects provided by humans. 
Differently, we leverage LLMs to automatically discover aspects to guide comparative reasoning, which provides a flexible way to incorporate fine-grained task-relevant signals while requiring minimal labeling efforts.

\section{Methods}

Our model, \model, produces comparative reasoning for text preference prediction that applies to densely written texts, generalizes to multiple domains, and ensures consistency. In this section, we give the generative process and inference procedure for \model. Our primary focus is ensuring the comparisons consistently distinguish similarities and differences between texts.

\subsection{Generative Process}
The generative process has three steps. First, given a text pair, \model simplifies the task by delineating a set of aspects, as depicted in Figure~\ref{fig:intro:fig}. These aspects, consisting of concise phrases,  enable the structured comparison between the texts. Second, \model produces concise comparisons, aspect-focused comparative statements that clearly express how the texts are similar and different.
In this paper, we implement this explicit consistency mechanism: similarities identified as shared between the text pair should not overlap with what's unique to each of them. 
Given the aspects and comparisons, the final step predicts which text is preferred.

Formally, for a text pair problem, we denote the text pair as $\leftarrow$ and $\rightarrow$, along with a query. 
\model has three components: Aspects $a = \{a_1, a_2, \ldots, a_n\}$, comparisons $c = \{c_1, c_2, \ldots, c_n\}$,
and text preferences \(y \in \{\leftarrow, \rightarrow\}\). 
The comparison $c$ has three columns: \(\{{c^{\leftarrow}}, {c_i^{\rightarrow}}, {c_i^{\cap}}\}, ~{c_i^{\rightarrow}} \text{and}~{c_i^{\leftarrow}}\) refers to properties exclusive to  $\rightarrow$ and $\leftarrow$ respectively, and \({c_i^{\cap}}\) to properties shared by both texts.

\model follows the following generative process:
First, it generates the aspects conditioned on the text using an \textit{aspect model}, $P(a)$. Second, comparisons for each aspect are generated from the comparison model 

$$P(c|a) \propto  \prod_{i}l(c_i)\times P(c_i|c_{<i}, a)$$

\noindent where the function $l:C \rightarrow \mathbb{R}^+$  evaluates the consistency of $c_i$. A higher value of $l(c_i)$ indicates a greater degree of consistency. 
Finally, preference model  $P(y|c, a)$ produces the preference label $y$.


\paragraph{Parameterization}
We use LLMs with specific prompts to parameterize each model. 
 With LLMs generating reliable scalar values of consistency is unreliable~\citep{imani2023mathprompter, liu2023improving}.  Instead of directly regressing a consistency score, we rely on pairwise comparisons, which have been observed to be more reliable \citep{qin2023large}. We define a pairwise comparator $l'(c, c') = \mathbbm{1}(l(c) \ge l(c'))$\footnote{We break the tie randomly.}, which takes a pair of comparisons $(c, c')$ and determines the more consistent one. 

To facilitate this, we recruit experts to develop few-shot prompts that demonstrate a direct comparison of two structured representations based on consistency within itself. We guide our annotators to assess pairs $(c, c')$ against consistency criteria, emphasizing that elements of the comparison should ideally exhibit no overlap. Detailed instructions are attached in the Appendix.

\begin{figure}[t]
    \centering        \includegraphics[width=1\linewidth]{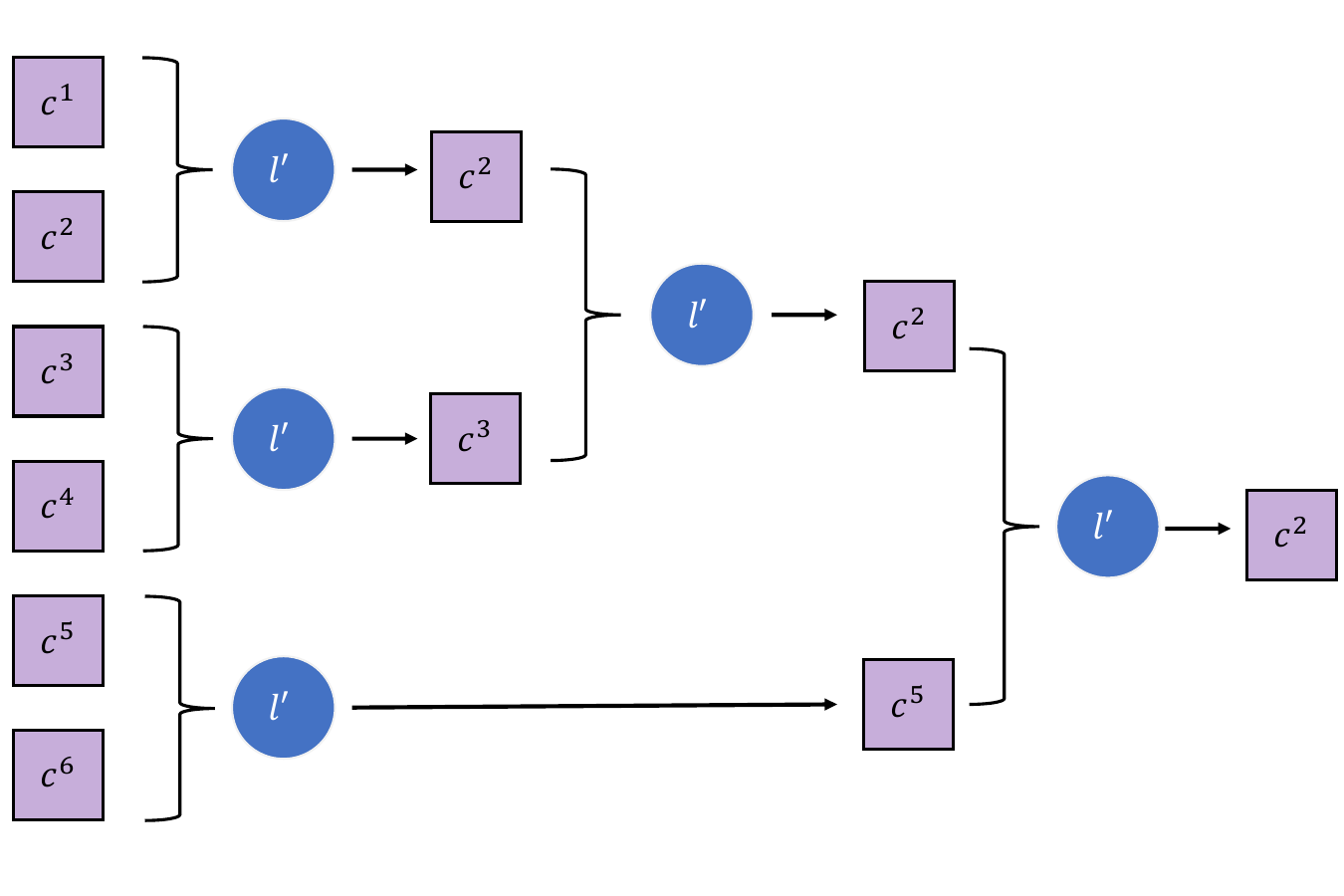}
\caption{Illustration of tournament inference. Given a set of samples, $C=\{c^1, c^2, c^3, c^4, c^5, c^6\}$, the tournament approach randomly partitions them into three groups in the first round, and each two is paired as input to $l'$ and output from $l'$ will be entering the next round. In this way, we only need to use $l'$ 5 times. }
    \label{fig:tournament}
\end{figure}

\subsection{Tournament-based Inference}
Given the generative model, the goal of inference is to produce aspects and comparisons that are high probability under the model and consistent.
We take a step-wise approach, choosing aspects, comparisons, and then finally predicting preferences.

When choosing aspects, we follow prior work by employing a variety of sampling strategies to obtain near-optimal aspects $a^*$ from $P(a)$ \citep{wang2022self, amplayo2021aspect}. We provide more details on these strategies in Section \ref{sec:exp:setup}.

Given aspects $a^*$, our next goal is to find comparisons that are likely under the comparison model $\argmax_c P(c|a^*) = \argmax_c l(c)\cdot P(c|a^*)$.
There are two challenges with this objective:
First, the set of possible comparisons is intractably large.
Second, the consistency function $l(c)$ is unreliable.
We approach the first challenge by sampling a set $C$ of high probability comparisons from $P(c|a)$, and the second challenge by selecting the most consistent comparison by applying the pairwise consistency comparator $l'(c,c')$ in a binary reduction. Formally, we select the most consistent comparison by optimizing
\[
c^{*} = \argmax_{c \in C} \sum_{c' \in C \setminus \{c\}} l'(c, c').
\] 
Naively, this optimization problem above requires $O(|C|^2)$ pairwise comparisons to optimize exactly. 

To reduce the number of pairwise comparisons, we utilize a tournament approach that performs $O(|C|)$ comparisons.
The tournament approach utilizes a binary reduction:
Each step of the binary reduction takes a pair of comparisons and eliminates the less logically consistent one into the successive rounds. We illustrate the tournament approach in Figure~\ref{fig:tournament}.
The naive and tournament approaches are equivalent if transitivity holds in the consistency comparator $l'(c,c')$. In practice, transitivity does not always hold with LLM parameterizations, resulting in the tournament approach trading off accuracy for efficiency.

Finally, with structured intermediate representation $(a^*,c^*)$, \model decides between $\leftarrow, \rightarrow$ which one is preferred by taking $\argmax P(y|a^*,c^*)$. 

\section{Experimental Setup}\label{sec:exp:setup}

\paragraph{Aspect Model} We experiment with two models for generating aspects: the online aspect model and the offline aspect model. Both models use PaLM-2-L to obtain aspects.

The online aspect model dynamically generates aspects using the CoT paradigm ~\citep{adlakha2023evaluating} to deduce aspects based on text inputs and applies self-consistency~\citep{wang2022self} to select the most agreeable aspect for each text pair. However, this model is costly due to the extensive use of LLM API calls for every pair of tasks.

The offline aspect model extracts aspects from a collection of text corpora, adapting the concept from \citet{pang2021agreesum} but employing LLMs. Specifically, this model prompts an LLM to extract aspects from each text within the collected corpora (50 pairs of texts for each task in this paper). It then prompts an LLM to refine and consolidate all generated aspects. Finally, we identify five fixed aspects as to use directly for any text pair of one task. This aspect model significantly reduces costs by allowing offline refinement of aspects. Refined aspects are fixed, thus they can be directly used without any additional expense.

In our experimental studies, we report only the best results for both baselines and in this section. To understand the impact of the aspect model, we detail its effects in our analysis section.

\paragraph{Comparison Model} We use PaLM-2-L as the major LLM backbone of comparison model of \model to produce intermediate structured representations.

\paragraph{Preference Model} For the final text preference prediction model, we experiment with two other LLM backbones differing in their model capacity. We aim to prove that the intermediate structured representations produced by \model with PaLM-2-L can help any backbone LLMs to predict text preference more accurately, regardless of their capacity. Specifically, we have used OpenAI's GPT-3.5, and GPT-4~\citep{gpt4} in our experiments.

\paragraph{Prompting Templates} Prompts used in different models can be found in our Appendix. Note that we do not tailor the preference model's prompts; instead, we adapted the templates from \citet{rafailov2023direct} for a fair comparison across baselines\footnote{In the original DPO paper~\citep{rafailov2023direct}, the authors did not use the Anthropic-Harmless dataset, we adapted their templated for Harmless datasets.}. 

\paragraph{Hyperparameters} As \model searches for the best comparisons during the inference stage, as a result, we have a hyperparameter \(|C|\), referring to the number of samples generated by the comparison model. \(|C|\) is an important parameter that might affect the quality of the intermediate structured representation produced by . For the reported results in this section, we set \(|C|=8\). We study the influence of this hyperparameter in Sec \ref{sec:analysis}.

\paragraph{Baselines} For evaluation, we consider several baselines, primarily focused on the LLM-based prompting approaches. Below is a detailed overview of these baselines:

\noindent (1) \textit{Direct Prompting (DP):} This method directly prompts LLMs to predict text preference.

\noindent (2) \textit{DP w/Aspects:} This approach is a variation of DP. The difference is that DP w/Aspects incorporates aspects generated by the aspect models.

\noindent (3) \textit{CoT-0-shot:} This baseline utilizes a standard CoT-0-shot template for task preference prediction (with "let's think step by step"). More details of the prompt template are available in the appendix.

\noindent (4) \textit{CoT-1-shot:} In addition to zero-shot prompting, we also carry out experiments using a 1-shot example within the CoT paradigm. For that purpose, we craft our 1-shot examples across different datasets.

\noindent (5) \textit{CoT-SelfCon:} This baseline integrates self-consistency to CoT-0-shot baseline aiming to remove the arbitrariness.

Specifically, CoT-SelfCon first samples multiple responses from an LLM using the same prompt and text pair input. Subsequently, CoT-SelfCon aggregates all responses to identify the most frequent answer. In our experimental studies, we set the number of sample responses to 8 and use a majority vote to determine the desired response, randomly selecting a response in the event of a tie.




\paragraph{Datasets}(1) \textbf{TL;DR~\citep{stiennon2020learning}}: We use OpenAI's filtered Reddit and CNN/Daily Mail TL;DR dataset. OpenAI also created a preference dataset from this, where labelers rated two generated summaries per post. For the CNN/Daily Mail part, for a given piece of news, we extracted two graded summaries and used the overall score to create the label. More details can be found in the original paper.

\noindent (2)~\textbf{RLAIF-HH~\citep{bai2022training}}: The RLAIF-HH from Anthropic dataset comprises dialogues from interactions between crowdworkers and large language models. In these exchanges, workers either seek assistance or provoke potentially harmful responses from the AI. The responses are then labeled based on their helpfulness or harmfulness. 

\noindent (3)~\textbf{TREC News~\citep{trecnews}:} The TREC News dataset contains 
query-document pairs focused on ad-hoc ranking and filtering tasks from the late 1980s to early 2000s. We modify the dataset as follows for preference prediction: for a given query, we extract two document answers to construct the triplet and use the relevance score provided by the original dataset to decide which document is more preferred.

\paragraph{Dataset Sampling} As datasets that have been used in the past are in large volumes, we only sampled a small ratio of them due to the cost of running all experiments. We sample roughly 250-300 data points from each dataset uniformly.  More details of the sampled dataset can be found in Table \ref{tab:dataset:statistics}
\begin{table}[tb]
\centering
\resizebox{0.92\linewidth}{!}{
\begin{tabular}{l|ccc} 
\toprule
Dataset& \# Samples & Avg. Length\\
\midrule
TL;DR-CNN/DM&  256& 572 \\
 TL;DR-Reddit& 259& 362 \\
Antropic-Helpful&  250& 102  \\
 Antropic-Harmless& 249& 93 \\
TREC News &  291&  947 \\
\midrule
AVG&  278& 433 \\ 
\bottomrule
\end{tabular}}
\caption{Statistics of Datasets in Experimental Studies}\label{tab:dataset:statistics}
\end{table}

\paragraph{Metrics} We report the accuracy of all approaches in our experiment  ($\frac{\text{Correctly Predicted Instances}}{\text{All Instances}}$) to measure the performance. 


\begin{table*}[tbh]
\centering
\begin{tabular}{cc|ccc|ccc|c} 
\toprule
Preference& Comparison& \multicolumn{3}{c|}{TLDR}& \multicolumn{3}{c|}{RLAIF}& Document \\
Model& Model& \multicolumn{3}{c|}{}& \multicolumn{3}{c|}{}& Ranking \\
&                             &          Reddit&             CNN/DM&          AVG&              Helpful  &    Harmless&       AVG &        TREC News\\
\toprule
 &  DP& 62.89& 61.39& 62.14& 58.40& 58.15& 58.27& 44.36\\
 & DP w/Aspects& 62.50& 62.55& 62.52& 59.20& 53.72& 56.46& 46.18\\ GPT-3.5&                             CoT-0-shot&          63.67&             64.48&          64.08&             59.00&              56.94&        57.97&                47.64\\
 &                             CoT-1-shot&          64.06&             63.71&          63.88&             59.20&              58.55&        58.88&                50.18\\ 
&                             CoT-SelfCon&          64.92&             63.32&          64.12&             60.60&              58.75&        59.68&                50.55\\ 
& 
\model &             \textbf{68.36}&             \textbf{68.34}&          \textbf{68.55}&             \textbf{63.20}&              \textbf{59.76}&        \textbf{61.49}&                \textbf{52.95}\\
\midrule
 & DP& 66.41& 64.86& 65.63& 62.60& 58.85& 60.58& 52.00\\
 & DP w/Aspects& 65.63& 65.25& 65.44& 60.60& 60.97& 60.78& 55.64\\  
GPT-4&                             CoT-0-shot&          68.75&             68.34&          68.54&             63.00&              60.56&        61.78&                59.64\\
 &                             CoT-1-shot&          69.92&             69.50&          69.71&             63.80&              60.16&        61.98&                61.09\\
 & CoT-SelfCon& 71.67& 69.12& 69.90& 64.00& 60.76& 62.38& 61.82\\
&   
\model &             \textbf{73.83}&             \textbf{70.65}&          \textbf{72.25}&             \textbf{66.60}&              \textbf{62.98}&        \textbf{64.79}&                \textbf{64.73}\\
\bottomrule
\end{tabular}
\caption{Experimental results of \model across different datasets in three different domains. DP refers to direct prompting. We use accuracy to measure the performance and report averaged the results from 5 rounds. }\label{tab:main:exp}
\end{table*}

\section{Results}
Experimental results in Table \ref{tab:main:exp} demonstrate \model's strong performance across all evaluation domains, with average gains of $\sim2.5$ and $\sim7.0$ points over the top and bottom baselines respectively. This confirms the benefits of structured comparative reasoning for enhanced text preference prediction. Using structured intermediate representations produced by \model, the preference prediction model better handles these comparative reasoning difficulties.

Moreover, we observe the input length as an additional factor impacting performance. For instance, the TREC News dataset comprises considerably longer texts than other corpora. Here, the DP method lags behind \model by over $9$ points, compared to the average $7$ point deficit across baselines. Though input length serves as an imperfect proxy for complexity, the results also signaled the potential benefit of using our method for longer inputs. 

We also want to point out that \model could be further improved by coupling with some of the existing general prompting techniques, for example, self-consistency~\cite{wang2022self} and self-verification~\cite{madaan2023self}. 


\section{Analysis}\label{sec:analysis}

To further understand the benefit of using \model to produce an intermediate structured representation, in this section, we conduct ablation studies and in-depth analysis. We also implement a user study to explore the potential of using \model to inform human beings' decisions. 
\begin{figure}[t]
    \centering        \includegraphics[width=.7\linewidth]{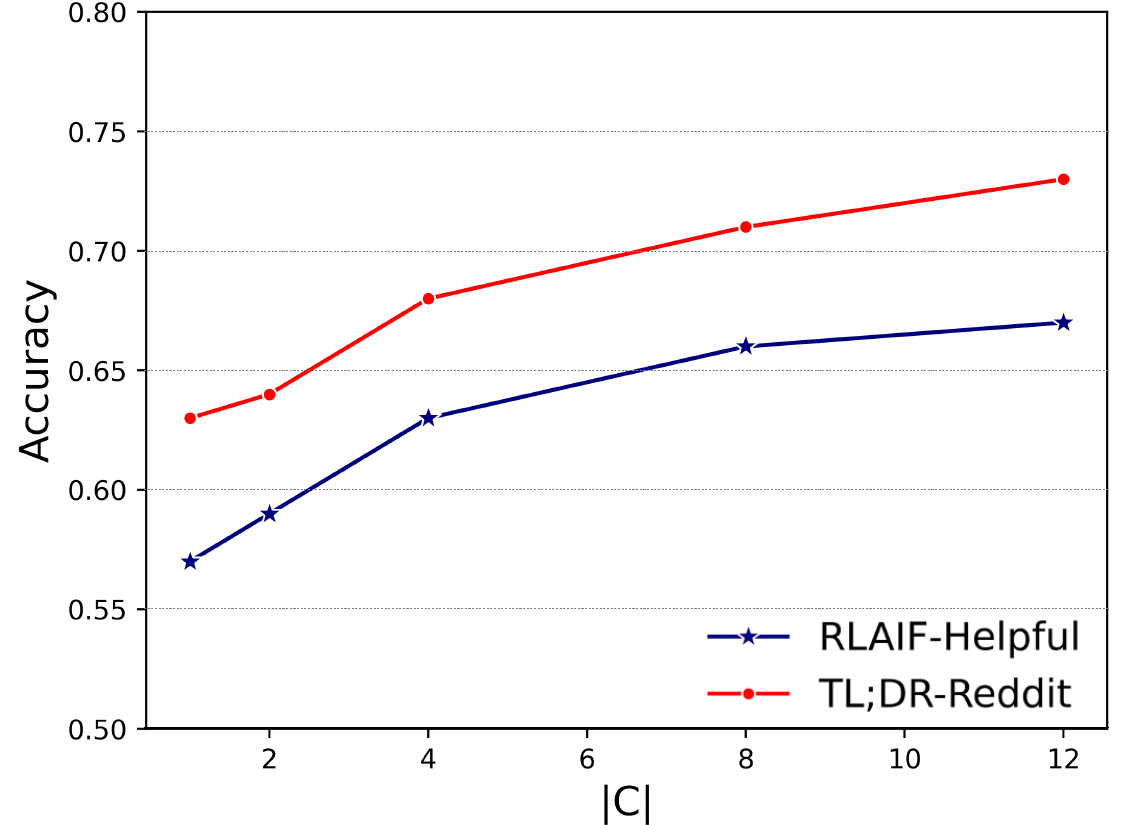}
\caption{Impact of \# samples $|C|$ in \model.} \vspace{-1em}
    \label{fig:analysis:c}
\end{figure}

\begin{table*}[t]
\centering
\begin{tabular}{ccc|cc|cc|c}
\toprule
Preference&Aspect&Model& \multicolumn{2}{c|}{TLDR}& \multicolumn{2}{c|}{RLAIF}& Document \\
Model&  Model  &&\multicolumn{2}{c|}{}& \multicolumn{2}{c|}{}& Ranking \\
&                 &&          Reddit &             CNN/DM&                   Helpful  &    Harmless&             TREC News\\
\toprule
 & Online &DP w/Aspects& 62.89 & 62.16 &  59.20 & 53.72 & 46.18\\
 & Online &\model& 67.97& 67.95 &  63.00 & 59.15 & \textbf{53.09}\\
GPT3.5  &  Offline &DP w/Aspects& 62.50 & 62.55 &  58.80 & 53.32 & 44.96 \\
 & Offline &\model& \textbf{68.36}& \textbf{68.34}&  \textbf{63.20}& \textbf{59.76}& \textbf{53.09}\\\midrule
& Online &DP w/Aspects& 65.63 & 63.32 & 60.40 & 60.97 & 54.81 \\

& Online &\model& 73.05 & 70.27 &  66.00 & 62.37 & 63.81 \\
GPT4 & Offline &DP w/Aspects& 64.84 & 65.25 & 60.60 & 59.76 & 55.64 \\ 
& Offline &\model& \textbf{73.83}& \textbf{70.65}&  \textbf{66.60}& \textbf{62.98}& \textbf{64.73}\\\toprule
\end{tabular}
\caption{Calibration of different aspect models. Online refers to Online Aspect Model, and Offline refers to Offline Aspect Model. DP w/Aspects refers to Direct Prompting with Aspects. }
\label{table:aspect:model:performance}
\end{table*}

\subsection{Effectiveness of Pairwise Comparator}

To calibrate the effective gain arising from the pairwise comparator $l'$, we first compare variants of \model with the comparators and those with different hyperparameter configurations of  \model. We use different intermediate structured representations produced by variants of \model to predict the text preference. Results are shown in Figure \ref{fig:analysis:c}.

With $|C|=1$, where there is effectively no pairwise comparator $l'$, the performance of the preference model was found to be comparable to baseline results shown in Table \ref{tab:main:exp}. This suggests that inconsistent structured representations could potentially degrade the performance of the preference model. An increase in accuracy was observed with larger values of $|C|$, indicating the benefits of pairwise comparator. However, this improvement plateaued when $|C|$ exceeded 8, hinting at a potential ceiling effect for our approach, irrespective of further increases in $|C|$.

\subsection{Impact of Different Aspect Models}
To understand the effect of different aspect models, we conduct ablation studies comparing the baseline that used aspects and \model with aspect models proposed in our experimental study. 

Table \ref{table:aspect:model:performance} presents the results. It shows that \model with the offline aspect model consistently outperforms or performs as well as \model with the online aspect model. However, for the DP w/Aspects baseline, neither the online nor the offline aspect model demonstrates superiority over the other.
This indicates that \model does not require online LLM calls which dynamically generate aspects and can effectively utilize the offline aspect model to obtain aspects for the given task at pretty low cost.

\subsection{Cost Analysis of \model and Few-shot CoT-SelfCon}
As discussed in our experimental study, CoT-SelfCon has no pairwise comparator components, resulting in lower LLM usage. On the other hand, in our primary experimental studies, we utilize PaLM-2-L to create intermediate structured representations and other LLM for preference prediction to avoid potential overfitting. In contrast, the CoT-SelfCon baseline consistently employs the same LLM (GPT-4) all the way.

\begin{table}[t]
\centering
\begin{tabular}{c|ccc}\toprule
 Total LLM calls&  8  & 15 & 24\\\midrule
 \model& \bf 0.682  & \bf 0.738& \bf 0.750\\
 CoT-SelfCon& 0.678& 0.728& 0.730\\\toprule
\end{tabular}
\caption{Accury of text preference prediction of \model against CoT-SelfCon with the same \#  of LLM calls.}\label{tab:cost2}
\end{table}

To ensure a fair comparison and eliminate biases that might arise from using different LLMs and \# total LLM calls, we only use GPT-4 for both \model and CoT-SelfCon in this analysis. We use a fixed number of total LLM calls, including the generation of intermediate structured representations and the final preference prediction. We limit our experiments to a single dataset with $100$ samples and average the results over 5 rounds for the cost consideration. The results are shown in Table \ref{tab:cost2}. Our analysis indicates that with the same \# total LLM calls and the same LLM backbone, \model predicts preference consistently more accurately.

Furthermore, we evaluate against few-shot CoT-SelfCon, commonly regarded as a strong baseline. Given that \model is in a zero-shot setting in our experiments, for a fair comparison, we compare few-shot \model with few-shot CoT-SelfCon, varying \# LLM calls and \# few-shot examples. We limit \# few-shot examples to 5. This makes sure the context length is within the LLM's length window.

Results are shown in Table \ref{tab:cost:few:shot}. When the total LLM calls are low, CoT-SelfCon maintains a slight advantage over \model. However, as the number of LLM calls increases, \model consistently outperforms few-shot CoT-SelfCon with the margin widening. This trend is attributed to the necessity for pairwise comparators to produce logically consistent intermediate-structured representations, leading to more accurate predictions.

\begin{table}[t]
\centering
\begin{tabular}{cc|cc}\toprule
 Total &  \# Few-shot & CoT-  & \model \\
 LLM calls& Examples & SelfCon & \\\midrule
 8& 3& \textbf{0.678} & 0.672\\
 & 5& \textbf{0.694} & 0.685\\\midrule
  15& 3& 0.718& \textbf{0.733}\\
 & 5& 0.742& \textbf{0.756}\\\midrule
  24& 3& 0.778& \textbf{0.797}\\
 & 5& 0.796& \textbf{0.812}\\\toprule
\end{tabular}
\caption{Accury of text preference prediction of few-shot \model against few-shot CoT-SelfCon with the same number of LLM calls.}\label{tab:cost:few:shot}
\end{table}


\subsection{Efficiency of Tournament Approach}

We study the efficiency and effectiveness of the tournament approach w.r.t. other inference methods. Random Selection refers to the process of randomly selecting one sample from $C$ during the inference stage, while Exact Search involves running all possible comparisons, which takes $O(n^2)$. We measure the cost using the total input length and the number of LLM calls, as this is common practice for the actual cost calculation in commercial Large Language Models (LLMs). We used the same dataset from the previous subsection. 

We find a significant gap between the Random Selection approach and the other two approaches as shown in Table \ref{tab:cost}. Although Exact Search yields the best results, it requires $4$ times the token length and $49$ more LLM calls, potentially leading to a substantial increase in cost.

\begin{table}[t]
\centering
\resizebox{0.95\linewidth}{!}{
\begin{tabular}{c|ccc}
\toprule
  &  Random&Tournament&  Exact\\
 &  Selection&Scheme& Search\\
\midrule
\# LLM calls&  1&7&  56\\ 
Decoded Len&  372&2,651&  13,272\\ 
Accuracy&  0.63 & 0.71&  0.73\\ 
\bottomrule
\end{tabular}}
\caption{Cost and accuracy analysis of different inference approach of \model.}\label{tab:cost}
\end{table}

\subsection{Human Evaluation}
We conduct additional human evaluations to see how the intermediate structured representations produced by \model inform human decision-making. 

\paragraph{Annoators}We recruit our annotators from an internal pool. Demographic and geographic characteristics of the annotator population are not accessible to our researchers. Information can be used to identify annotators that are fully anonymized. Consent forms have to be signed by annotators to take part in this study. 

\paragraph{Study Design}In consideration of ethical standards and the requirement to avoid directly testing annotators, we structure our human evaluation as follows: Annotators are presented with a query alongside a pair of text options, denoted as $(\leftarrow, \rightarrow)$. They determine which text, either $\leftarrow$ or $\rightarrow$, is preferable. They have three options: $\leftarrow$ is better, $\rightarrow$ is better, and tie. Following their initial decision, annotators are then shown the intermediate structured representations generated by different variants of \model. They decide if this additional information leads them to reconsider their initial choice and provide reasons for any change in their decision. This evaluation process uses two variants of \model: $|C|=1$ and $|C|=8$ respectively. For ethical considerations, we only experiment with RLAIF-helpfulness and TL;DR-Reddit, ensuring the content is not harmful or violent manually. We instantiate $100$ data points for each dataset and assign each question to three annotators. We collect $96$ and $98$ questions with useful responses from all three annotators for RLAIF-helpful and TL;DR-Reddit respectively.

\begin{figure}[t]
    \centering        \includegraphics[width=1\linewidth]{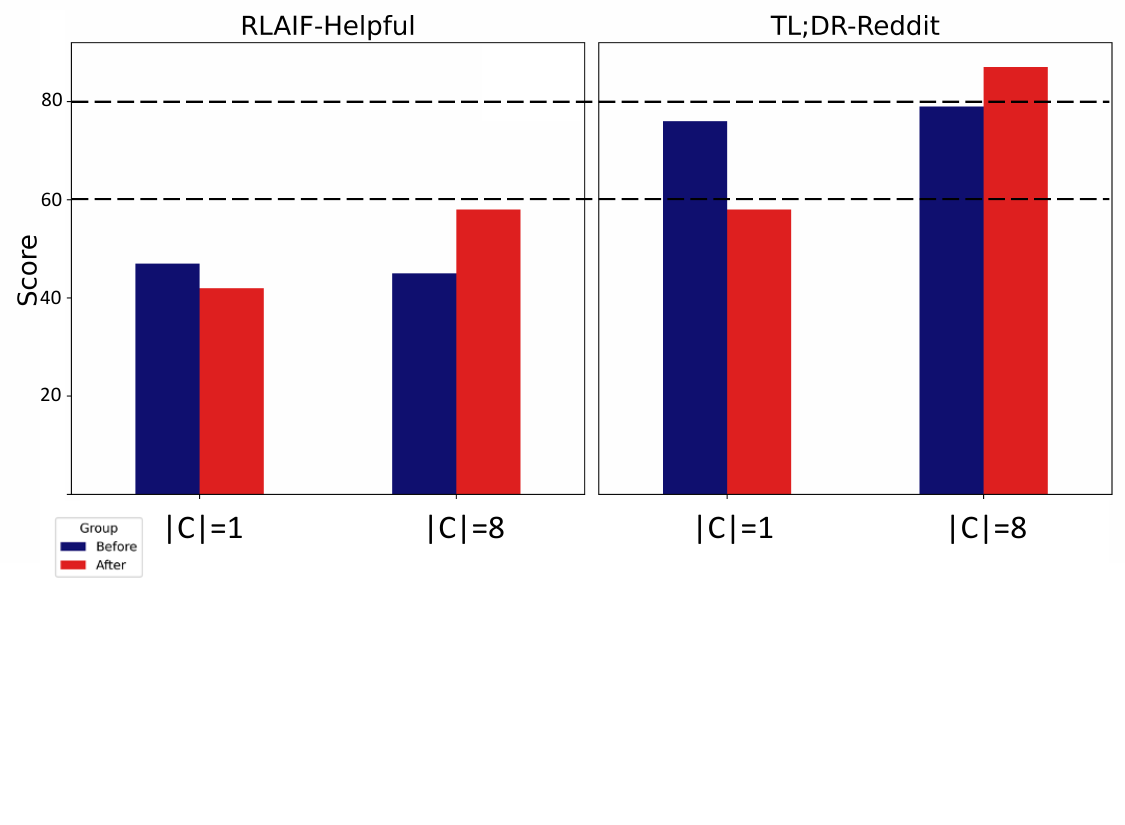}\vspace{-5em}
\caption{Human evaluation on structured representation produced by different settings of \model.} 
    \label{fig:human:eval}
\end{figure}

\paragraph{Metrics} We use the ground truth to gather the scores: we assign $1$ for any correct answer, $0$ for a \texttt{tie}, and $-1$ for any other incorrect answers. 

\paragraph{Findings} As shown in Figure \ref{fig:human:eval}, with the aid of more consistent intermediate structured representations ($|C|=8$), annotators are inclined to revise and flip their previous wrong answers to correct ones. This suggests that the intermediate structured representation may facilitate better decision-making among human evaluators. However, we also observe that intermediate structured representations without using a pairwise comparator ($|C|=1$) could mislead annotators, deterring them from selecting the correct preference. This amplifies the importance of the pairwise comparator to ensure consistency. 

We also look into quantitative justifications provided by annotators. Most annotators stated that intermediate structured representations helped them better understand two texts. One mentioned, \texttt{"the table gives the concise comparison"}, while another pointed out, \texttt{"this [table] helped me to understand better the implications of the two answers, and I changed my mind after reading [the table]"}. 
Besides, we also observe complaints about the intermediate structured representations being hallucinatory and not factual. 
The issue is more noticeable in cases where the structured representation is produced by \model without a pairwise comparator. 
This suggests that enforcing a pairwise comparator might mitigate the arbitrariness of LLM’s output for better consistency, but still poses the risk of presenting hallucinated results to annoators.
\section{Conclusion}
This paper presents \model, a structured comparative reasoning model for improving text preference prediction. \model constructs intermediate structured representations to explicitly contrast text pairs, incorporating a consistency comparator to enhance accuracy and coherence. Comprehensive experiments across text summarization, retrieval, and response rating tasks demonstrated that \model significantly improves consistency and achieves state-of-the-art performance. Analyses confirm the effectiveness of \model's structured reasoning approach and consistency enforcement. Our human evaluations show that \model interpretations can assist users in making informed decisions.
\section{Limitations}
This work has several limitations that provide opportunities for future investigation. First, the evaluation was conducted on a sample set of datasets that, while spanning diverse domains, might not fully characterize the breadth of real-world textual comparison needs. Expanding \model's testing to larger, multilingual corpora is essential to assess its full potential and limitations beyond English. 
Furthermore, there are likely upper bounds on \model's effectiveness imposed by the reasoning capacity of the underlying language model backbone. 
As more advanced LLMs emerge, exploring their integration could help quantify this ceiling effect. On a technical level, in this paper, measuring consistency relies on approximate metrics, so developing more rigorous evaluation schemes could better highlight \model's benefits. We also do not include other prompting techniques that have been well-studied in the community, which we leave for future work. 

\section{Ethical Considerations}
This research paper might risk potential biases that could arise from textual comparisons, particularly around sensitive attributes. \model is trained on established corpora like Wikipedia and books that may inherently contain societal biases. While a full analysis of these biases is beyond the scope here, we acknowledge the risk that \model may inherit problematic biases from its training data. Applying recent advancements in language bias detection to \model could help quantify and mitigate these risks. We are interested in exploring this as part of future work. Furthermore, this research focused solely on English; extending to other languages is an important direction that would require non-trivial adaptation. Overall, while showing promise, \model has significant scope for improvement as limitations around evaluation, multilingual capabilities, consistency measurement, bias, and applied usage are addressed through future work.

\section*{Acknowledgement}
We thank Pengcheng Yin, Ethan Liang, Wenting Zhao, Celine Lee, Woojeong Kim, Jack Morris, and Junxiong Wang for their valuable suggestions and feedback. J.NY and AMR are supported by NSF CAREER \#2037519, NSF III:\#1901030, and NSF \#2229873.

\bibliographystyle{acl_natbib}
\bibliography{main}

\section*{Appendix}

\begin{figure*}[t]
    \begin{mdframed}
    \textbf{Which of the following summaries does a better job of summarizing the most important points in the given forum article, 
    without including unimportant or irrelevant details?} \\

    A good summary is both precise and concise. \\

    \textbf{Original Article:} \\
    \{article\}\\

    \textbf{Summary A:} \\
    \{contextA\}\\

    \textbf{Summary B:} \\
    \{contextB\}\\

    Take a deep breath and think about this question step by step! FIRST, think step by step to have a comparison of the two summaries, explaining which you prefer and why. SECOND, on a new line, state only "A" or "B" to indicate your choice.\\

    Your response should use the format: \\
    Comparison: <step by step comparison> \\
    Preferred: <"A" or "B">.
    \end{mdframed}
    \caption{Preference model prompt for CoT Zero-shot Prompting for TL;DR}
\end{figure*}

\begin{figure*}[t]
    \begin{mdframed}
    \textbf{Which of the following summaries does a better job of summarizing the most important points in the given forum article, 
    without including unimportant or irrelevant details?} \\

    A good summary is both precise and concise. \\

    \textbf{Original Article:} \\
    \{article\}\\

    \textbf{Summary A:} \\
    \{contextA\}\\

    \textbf{Summary B:} \\
    \{contextB\}\\

     FIRST, explaining which you prefer and why. SECOND, on a new line, state only "A" or "B" to indicate your choice.\\

    Your response should use the format: \\
    Comparison: <step by step comparison> \\
    Preferred: <"A" or "B">.
    \end{mdframed}
    \caption{Preference model prompt for Direct Prompting for TL;DR}
\end{figure*}

\begin{figure*}[t]
    \begin{mdframed}
    \textbf{Which of the following summaries does a better job of summarizing the most important points in the given forum article, 
    without including unimportant or irrelevant details? You are also given some aspects to help you make the decision} \\

    A good summary is both precise and concise. \\

    \textbf{Original Article:} \\
    \{article\}\\

    \textbf{Summary A:} \\
    \{contextA\}\\

    \textbf{Summary B:} \\
    \{contextB\}\\

    \textbf{Aspects:} \\
    \{aspects\}\\

     FIRST, explaining which you prefer and why. In your evaluation, you need to consider aspects that are given above. SECOND, on a new line, state only "A" or "B" to indicate your choice.\\

    Your response should use the format: \\
    Comparison: <step by step comparison> \\
    Preferred: <"A" or "B">.
    \end{mdframed}
    \caption{Preference model prompt for Direct Prompting with Aspects for TL;DR}
\end{figure*}

\begin{figure*}[t]
    \begin{mdframed}
    \textbf{Which of the following summaries does a better job of summarizing the most important points in the given forum article, without including unimportant or irrelevant details? You are also given a comparative reasoning table that analyzes the differences and similarities between the two summaries. } \\

    A good summary is both precise and concise. \\

    \textbf{Original Article:} \\
    \{article\}\\

    \textbf{Summary A:} \\
    \{contextA\}\\

    \textbf{Summary B:} \\
    \{contextB\}\\

    \textbf{Comparative Reasoning Table:} \\
    \{table\}\\

     FIRST, explain which you prefer and why. In your evaluation, you can use the comparative reasoning table above to help you make the justification and the decision. SECOND, on a new line, state only "A" or "B" to indicate your choice.\\

    Your response should use the format: \\
    Comparison: <step by step comparison> \\
    Preferred: <"A" or "B">.
    \end{mdframed}
    \caption{Preference model prompt for \model for TL;DR}
\end{figure*}

\begin{figure*}[t]
    \begin{mdframed}
   \textbf{Which of the following documents aligns better with the query given?} \\

A good retrieved document should be relevant to the query. \\

\textbf{Query:} \\
\{query\}\\

\textbf{Document A:} \\
\{contextA\}\\

\textbf{Document B:} \\
\{contextB\}\\

Take a deep breath and think about this question step by step! FIRST, think step by step to have a comparison of the two retrieved documents, explaining which you prefer and why. SECOND, on a new line, state only "A" or "B" to indicate your choice.\\

Your response should use the format: \\
Comparison: <step by step comparison> \\
Preferred: <"A" or "B">.

    \end{mdframed}
    \caption{Preference model prompt for Zero-shot CoT Prompting for TREC News}
\end{figure*}

\begin{figure*}[t]
    \begin{mdframed}

\textbf{Which of the following documents aligns better with the query given?} \\

A good retrieved document should be relevant to the query. \\

\textbf{Query:} \\
\{query\}\\

\textbf{Document A:} \\
\{contextA\}\\

\textbf{Document B:} \\
\{contextB\}\\

FIRST, have a comparison of the two retrieved documents, explaining which you prefer and why. SECOND, on a new line, state only "A" or "B" to indicate your choice.\\

Your response should use the format: \\
Comparison: <step by step comparison> \\
Preferred: <"A" or "B">.

    \end{mdframed}
    \caption{Preference model prompt for Direct Prompting for TREC News}
\end{figure*}

\begin{figure*}[t]
    \begin{mdframed}

\textbf{Which of the following documents aligns better with the query given? You are also given some aspects to help you make the decision} \\

A good retrieved document should be relevant to the query. \\

\textbf{Query:} \\
\{query\}\\

\textbf{Document A:} \\
\{contextA\}\\

\textbf{Document B:} \\
\{contextB\}\\

\textbf{Aspects:} \\
\{aspects\}\\

FIRST, have a comparison of two retrieved documents, explaining which you prefer and why. In your evaluation, you need to consider aspects that are given above. SECOND, on a new line, state only "A" or "B" to indicate your choice.\\

Your response should use the format: \\
Comparison: <step by step comparison> \\
Preferred: <"A" or "B">.

    \end{mdframed}
    \caption{Preference model prompt for Direct Prompting with Aspects for TREC News}
\end{figure*}

\begin{figure*}[t] 
    \begin{mdframed}
    \textbf{Which of the following documents aligns better with the query given? You are also given a comparative reasoning table that analyzes the differences and similarities between the two documents. } \\

    A good retrieved document should be relevant to the query. \\

    \textbf{Query:} \\
    \{query\}\\
    
    \textbf{Document A:} \\
    \{contextA\}\\
    
    \textbf{Document B:} \\
    \{contextB\}\\

    \textbf{Comparative Reasoning Table:} \\
    \{table\}\\

     FIRST, explaining which you prefer and why. In your evaluation, you can use the comparative reasoning table above to help you make the justifications and decision. SECOND, on a new line, state only "A" or "B" to indicate your choice.\\

    Your response should use the format: \\
    Comparison: <step by step comparison> \\
    Preferred: <"A" or "B">.
    \end{mdframed}
    \caption{Preference model prompt for \model for TREC News}
\end{figure*}

\begin{figure*}[t] 
    \begin{mdframed}
   \textbf{For the following query to a chatbot, which response is more helpful?} \\

\textbf{Query to a Chatbot:} \\
\{article\}\\

\textbf{Response A:} \\
\{contextA\}\\

\textbf{Response B:} \\
\{contextB\}\\

Take a deep breath and think about this question step by step! FIRST, think step by step to have a comparison of the two responses generated, explaining which you prefer and why. SECOND, on a new line, state only "A" or "B" to indicate your choice.\\

Your response should use the format: \\
Comparison: <step by step comparison> \\
Preferred: <"A" or "B">.

    \end{mdframed}
    \caption{Preference model prompt for Zero-shot CoT Prompting for RLAIF-HH }
\end{figure*}

\begin{figure*}[t] 
 \begin{mdframed}

\textbf{For the following query to a chatbot, which response is more helpful?} \\

\textbf{Query to a Chatbot:} \\
\{article\}\\

\textbf{Response A:} \\
\{contextA\}\\

\textbf{Response B:} \\
\{contextB\}\\

FIRST, have a comparison of the two generated responses, explaining which you prefer and why. SECOND, on a new line, state only "A" or "B" to indicate your choice.\\

Your response should use the format: \\
Comparison: <step by step comparison> \\
Preferred: <"A" or "B">.

    \end{mdframed}
    \caption{Preference model prompt for Direct Prompting  for RLAIF-HH}
\end{figure*}

\begin{figure*}[t]
\begin{mdframed}

\textbf{For the following query to a chatbot, which response is more helpful? You are also given some aspects to help you make the decision} \\

\textbf{Query to a Chatbot:} \\
\{article\}\\

\textbf{Response A:} \\
\{contextA\}\\

\textbf{Response B:} \\
\{contextB\}\\

\textbf{Aspects:} \\
\{aspect\}\\

FIRST, have a comparison of the two generated responses, explaining which you prefer and why. In your evaluation, you need to consider aspects that are given above. SECOND, on a new line, state only "A" or "B" to indicate your choice.\\

Your response should use the format: \\
Comparison: <step by step comparison> \\
Preferred: <"A" or "B">.

    \end{mdframed}
    \caption{Preference model prompt for Direct Prompting with Aspects  for RLAIF-HH}
\end{figure*}

\begin{figure*}[t] 
 \begin{mdframed}
    \textbf{For the following query to a chatbot, which response is more helpful? You are also given a comparative reasoning table that analyzes the differences and similarities between the two generated responses. } \\

  \textbf{Query to a Chatbot:} \\
\{article\}\\

\textbf{Response A:} \\
\{contextA\}\\

\textbf{Response B:} \\
\{contextB\}\\

   \textbf{Comparative Reasoning Table:} \\
    \{table\}\\

     FIRST, explain which you prefer and why. In your evaluation, you can use the comparative reasoning table above to help you make the justifications and decisions. SECOND, on a new line, state only "A" or "B" to indicate your choice.\\

    Your response should use the format: \\
    Comparison: <step by step comparison> \\
    Preferred: <"A" or "B">.
    \end{mdframed}
    \caption{Preference model prompt for \model  for RLAIF-HH}
\end{figure*}

\begin{figure*}[t] 
\begin{mdframed}
\textbf{Which of the following summaries does a better job of summarizing the most important points in the given forum article, 
without including unimportant or irrelevant details? You are also given some aspects to help you make the decision} 

A good summary is both precise and concise. 

\textbf{Example Article:} \\
\{article\}\\

\textbf{Example Summary A:} \\
\{contextA\}\\

\textbf{Example Summary B:} \\
\{contextB\}\\

\textbf{Example Aspects:} \\
\{aspects\}\\

FIRST, explain which you prefer and why. In your evaluation, you need to consider aspects that are given above. SECOND, on a new line, state only "A" or "B" to indicate your choice.\\

Your response should use the format: \\
Comparison: <step by step comparison> \\
Preferred: <"A" or "B">.

Example Answer: 
\{example answer\}\\

Now, Based on the example above, take a deep breath and think about this question step by step to answer the following question. 

\textbf{Which of the following summaries does a better job of summarizing the most important points in the given forum article, 
without including unimportant or irrelevant details? You are also given some aspects to help you make the decision} \\

A good summary is both precise and concise. \\

\textbf{Original Article:} \\
\{article\}\\

\textbf{Summary A:} \\
\{contextA\}\\

\textbf{Summary B:} \\
\{contextB\}\\

\textbf{Aspects:} \\
\{aspects\}\\

FIRST, explain which you prefer and why. In your evaluation, you need to consider aspects that are given above. SECOND, on a new line, state only "A" or "B" to indicate your choice.\\

Your response should use the format: \\
Comparison: <step by step comparison> \\
Preferred: <"A" or "B">.
    \end{mdframed}
    \caption{Preference Model Prompts for CoT-1. }
\end{figure*}

\begin{figure*}[t] 
\begin{mdframed}
\textbf{Instructions:} Your task is to conduct a consistency analysis of two generated comparative table responses. Your evaluation should focus solely on the consistency of the responses. Each comparative table is constructed to delineate similarities and differences about a given query, juxtaposing candidate Summary 1 against candidate Summary 2. Consistency, in this context, refers to the logical coherence within each table. Specifically, for each row corresponding to an aspect-level comparison, the entries of the three columns that denote similarities should be distinct and non-overlapping with the entries that denote differences. A consistent response will differentiate between the commonalities and disparities, ensuring that the information under the 'similarities' column does not overlap with what is presented under the 'differences' column. This clear segregation is crucial in assessing the quality of the responses and their effectiveness in summarizing and contrasting the key points from the summaries.\\

\textbf{Query to a Chatbot:} \\
\{article\}\\

\textbf{Summary 1:} \\
\{contextA\}\\

\textbf{Summary 2:} \\
\{contextB\}\\

\textbf{Comparartive Table Response A:} \\
\{contextA\}\\

\textbf{Comparartive Table Response B:} \\
\{contextB\}\\

More consistent: <"A" or "B">. \\
Justifications: <Justifications>.

    \end{mdframed}
    \caption{Instructions to Craft prompts for Pairwise Comparator.  }
\end{figure*}

\end{document}